\documentclass[11pt]{article}

\usepackage[final]{acl}
\usepackage{graphicx}
\usepackage{booktabs}
\usepackage{enumitem}
\usepackage{pdflscape}
\usepackage{url}
\graphicspath{ {./images/} }
\usepackage{booktabs}
\usepackage{tcolorbox}

\usepackage{times}
\usepackage{latexsym}

\usepackage[T1]{fontenc}

\usepackage[utf8]{inputenc}

\usepackage{microtype}

\usepackage{inconsolata}

\usepackage{subcaption}

\usepackage{fancyhdr}

\title{Effects of Varying LLM Access on Essay Writing Behavior}

\author{Julia Christenson, Karin de Langis, Shirley Anugrah Hayati, Dongyeop Kang\\ \\
University of Minnesota \\ \\ \{chri5306, dento019, hayat023, dongyeop\}@umn.edu}

\begin{document}


\maketitle

\begin{abstract}

Investigating the degree to which large language models (LLMs) affect teaching and learning in universities can help identify strategies for integrating LLMs in a way that supports, rather than undermines, student learning outcomes. This study examined how varying levels of LLM assistance affect writing performance, engagement, and perceived authorship. 
We report a pilot study in which 24 college students were randomly assigned to write a short essay with no LLM access, limited access ($<=$3 prompts, responses capped at 100 words), or unlimited access. Overall essay quality was statistically indistinguishable across groups. 
Yet writing behavior and perceived authorship diverged sharply: students with limited access reported higher ownership (62.5\% would submit the essay as independent work, vs. 25\% in the unlimited group), stronger organizational gains, and more strategic, revision-focused prompting. 
The unlimited group spent more time writing, produced essays more similar to LLM output, and reported reduced creative expression. Our findings suggest that constraining, rather than banning, LLM access may preserve authorship confidence while retaining the scaffolding benefits of AI assistance.

\end{abstract}

\section{Introduction}
LLMs have become increasingly embedded in academic writing practices, reshaping how students plan, compose, and revise text. LLM-based tools such as ChatGPT and Grammarly now offer real-time feedback and idea generation, lowering barriers to writing while simultaneously raising questions about authorship, learning, and skill development. As these systems become normalized in educational contexts, it is important to understand their influence on student writing behavior. 

Prior work primarily frames LLM-assisted writing as a productivity and accessibility aid, reporting gains in fluency, error reduction, and brainstorming support across literacy levels, as well as tools that scaffold multiple stages of the writing process \cite{weijers-etal-2024-quantifying, lee2024design, Zhang2025college, xu2025patterns}. At the same time, researchers caution that sustained reliance on LLM may undermine independent writing, creativity, and critical thinking \cite{Rohilla2025AIOveruse, gerlich2025}. Others find no significant effect on overall academic achievement \cite{Yoo2025AIWriting}. Students often report reduced confidence and creativity with prolonged use, and prior work also highlights risks related to dependency, ethical ambiguity, and homogenized or biased outputs \cite{Zhai2024AIOverreliance, Chakra2025homogen}. Human-AI interaction research further shows that AI involvement can negatively influence perceptions of authorship and writing quality, particularly when disclosed \cite{10.1145/3637875, li-etal-2024-disclosure, Mansour2024selfdisc, Joshi_2025, gerlich-structure-2025}.

Existing research has shown that the use of LLMs has negative cognitive effects \cite{kosmyna2025brainonchat}, but we hypothesize that the degree of LLM access may play an important role in the efficacy of LLM assistants in educational contexts. 
This study addresses the gap by examining how different levels of LLM assistance influence students' writing by formulating three main research questions:

\begin{itemize}
    \item \textbf{RQ1}: How does varied LLM assistance influence students’ writing behavior, including time allocation, writing strategies, and prompting practices?
    \item \textbf{RQ2}: How do students perceive authorship, ownership, and acceptability of LLM-assisted writing?
    \item \textbf{RQ3}: How does the level of LLM assistance affect writing quality, originality, and lexical diversity in student essays? 
\end{itemize}

To address these questions, we conduct a study in which twenty-four college students wrote an essay in response to a prompt. By comparing writing produced under different levels of LLM assistance, this work aims to clarify when LLM assistance enhances writing efficiency and when it begins to erode core literacy skills.

We found that both LLM-assisted groups engaged in similar write-prompt-edit cycles, but with different strategic orientations: limited-access participants focused prompts on major revisions and argument development, while unlimited-access participants concentrated on minor mechanical edits. These behavioral differences mapped onto perceived authorship: limited-access participants reported stronger feelings of ownership, greater organizational improvement, and higher willingness to submit the essay as their own work. Essay grades and lexical diversity were marginally higher in the unlimited group, though differences were not statistically significant. Taken together, the findings suggest that constraining LLM access, rather than eliminating it, may offer a middle path between the cognitive risks of over-reliance and the scaffolding benefits of AI assistance.

\section{Related Work}

Prior research has examined human-AI interactions for scholastic writing needs through engagement with LLMs and behavioral outcomes. Several studies on college students' use of LLMs find that learners primarily rely on LLMs for brainstorming, revision, and information retrieval \cite{jelson2025chatgptessay, kondoro-2025-ai, WANG2024100247}. Although LLM assistants effectively correct surface-level language errors, they have been shown to reduce syntactic complexity and linguistic diversity with sustained use \cite{wang-spitz-2025-quantifying, Bui2025gen, Pan2025collab}. Some studies found that students with higher AI literacy and more strategic prompting practices tend to produce stronger writing, while students with lower performance often under-utilize or misuse AI tools \cite{Nguyen03052024, article, Joshi_2025}. A few studies observed modest improvements in students' independent learning and English writing ability when such tools are used alongside instruction \cite{10.1108/LHT-05-2020-0113, seo2024exploring, Xiao2024esl, Shen2025esl}. However, these prior works share limitations. Most studies employ binary comparisons—AI versus no-AI—rather than examining a response relationship to varied levels of AI access. This framing obscures whether the negative effects of LLM use are properties of AI assistance itself or the degree to which it is provided.

Growing concerns about dependence on LLMs negatively affecting critical thinking skills are prevalent. Survey research reveal many students and teachers are aware of the over-reliance on LLM writing assistants and concerns over academic dishonesty \cite{aliakbari2026aiusage, Joshi_2025, kondoro-2025-ai, khalid-2025-pakistan}. Several studies show lowered critical thinking abilities for those who exhibit higher dependence on LLMs \cite{kosmyna2025brainonchat, gerlich2025}. 

Recent studies pivoted to researching methods to increase perceived ownership and transparency through tailored LLM assistants. One study has found that greater AI transparency and interpretability are associated with more positive student experiences and reduced misuse, including academic dishonesty and diminished learning outcomes \cite{CUI2025100391}. Several works have studied human-AI collaboration patterns and suggest writer-centered approaches to LLM-based writing assistants, \cite{mysore-etal-2025-prototypical, guo2024preserving, liu-2025-writingcenter, gerlich-structure-2025, shibani-2024-shallow}, but largely emphasize interaction strategies and design recommendations without measuring whether such interventions actually improve writing outcomes or reduce problematic dependency patterns. 

Empirical studies demonstrate the benefits of LLM assistance, and the detrimental effects of AI on cognition and critical thinking skills, but the impact in real-world academic settings remains unclear \cite{Yoo2025AIWriting}. Together, these gaps motivate our study focused on how constraining LLM access can impact how students navigate a real-world academic situation. 

\section{Approach}

We aimed to investigate the extent to which students' writing experiences are impacted by the assistance LLM provides for them. The primary concern of the study was whether limiting LLM use provides a viable middle ground between the negative cognitive effects of LLM use and the benefits.\footnote{The study was approved by the IRB.}

\begin{table}[t]
\centering
\small
\begin{tabular}{ll}
\toprule
\textbf{Category} & \textbf{Percentage} \\
\midrule
\multicolumn{2}{l}{\textbf{Race}} \\
\midrule
White & 41.7\% \\
Asian & 16.6\% \\
Black or African-American & 8.3\% \\
Hispanic & 4.2\% \\
Two or more races & 25.0\% \\
Prefer not to disclose & 4.2\% \\

\hline
\multicolumn{2}{l}{\textbf{Gender}} \\
\midrule
Female & 75.0\% \\
Male & 25.0\% \\

\midrule
\multicolumn{2}{l}{\textbf{Major}} \\
\midrule
STEM & 41.7\% \\
Health and Medicine & 29.2\% \\
Arts and Humanities & 16.7\% \\
Business & 8.3\% \\
Social Sciences & 4.2\% \\

\midrule
\multicolumn{2}{l}{\textbf{Academic Level}} \\
\midrule
Senior & 66.7\% \\
Junior & 16.7\% \\
Sophomore & 4.2\% \\
Freshman & 8.3\% \\
Graduate & 4.2\% \\

\bottomrule
\end{tabular}
\caption{Participant demographics (N=24).}
\label{tab:demographics}
\end{table}
\subsection{Participants}
We recruited a total of 24 undergraduate and graduate students with 18 different majors through email recruitment. Details about participant demographic information are provided in Table \ref{tab:demographics}. 71\% of the participants rated themselves as strong writers by selecting either a 4 or 5 on a Likert scale out of 5. Participants, on average, have used LLM-based tools in some way in their previous 5 out of 10 writing assignments. Most students who use LLM tools in coursework reported using ChatGPT and built-in tools such as Grammarly. Participants who regularly use LLM-based tools typically employ them to generate ideas, seek assistance when they get stuck, and understand the material better. The participants who used LLM tools less frequently described the writing output as repetitive and superficial, and noted that they avoid LLM tools due to environmental concerns and academic integrity.

\subsection{Study Procedure}
First, participants completed a preliminary survey to determine eligibility. The eligibility requirements were native-level English fluency and identifying as an undergraduate or graduate student. Participants were randomly placed into one of three groups: control, limited access to LLM, or unlimited access. Participants then completed an approximately 30-minute in-person writing session. All participants provided informed consent under an IRB-approved protocol. Participants were given instructions to respond to an essay prompt in at least 300 words, using a text document. Instructions can be found in Appendix \ref{sec:Appendix_procedure}.
Each participant received the same prompt, shown below:

\begin{figure}
\centering
\includegraphics[width=\columnwidth, trim=.5cm .5cm 0.5cm 0.5cm, clip]{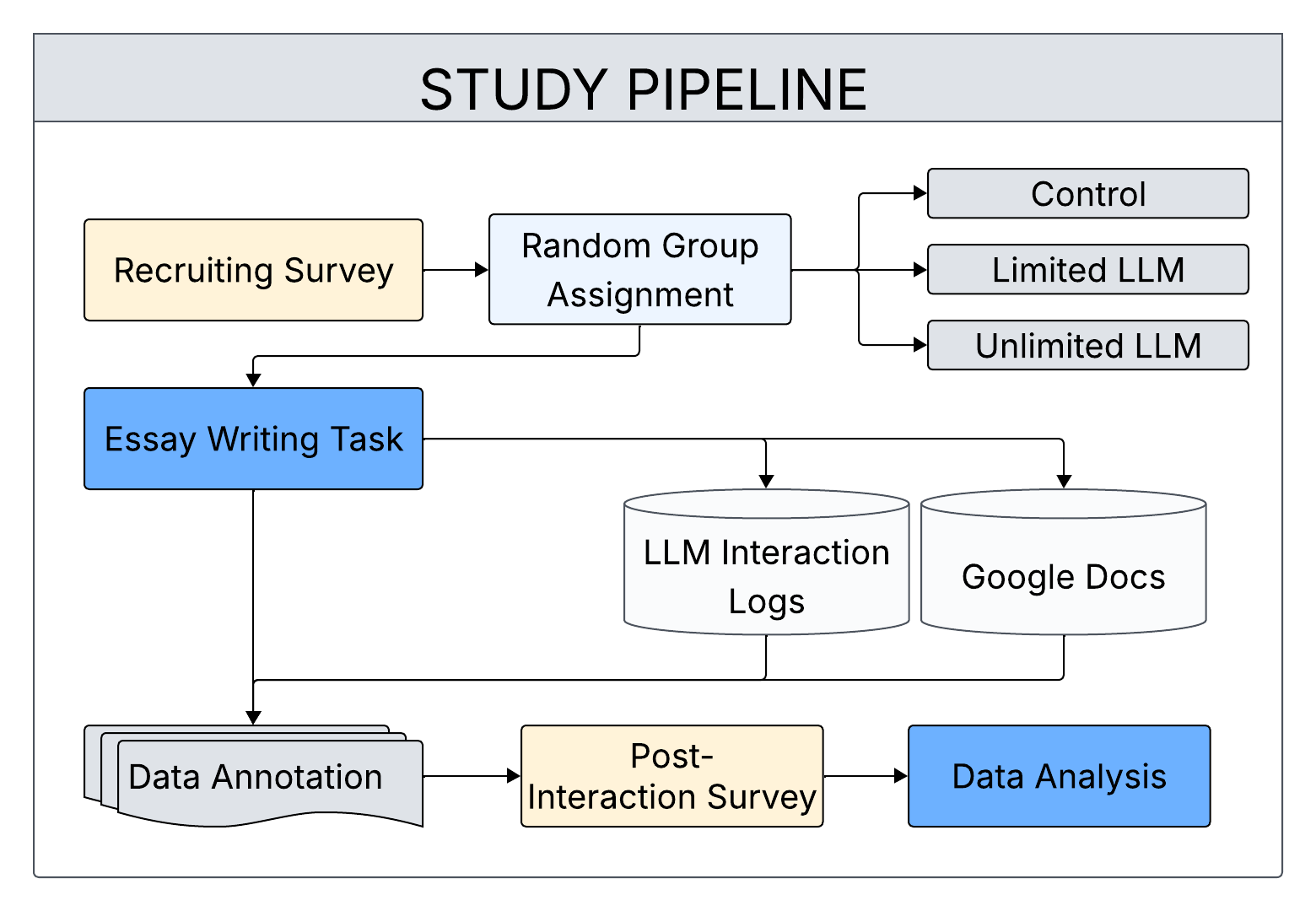}
\caption{Study pipeline.}
\label{pipeline}
\end{figure}

\begin{tcolorbox}[colback=gray!5, colframe=black, title=Essay Prompt]
Write a response in which you discuss the extent to which you agree or disagree with the recommendation and explain your reasoning for the position you take.
\newline \newline
College students should be encouraged to pursue subjects that interest them rather than the courses that seem most likely to lead to jobs.
\end{tcolorbox}


Since participants completed the writing task on their personal computers, we also instructed them to disable all AI writing extensions (e.g., Grammarly) on their browsers. 

Participants in the control group did not receive any help from LLM when writing their essays, while participants in other groups have access to chat with LLM-based chatbots during their writing process. To access the LLM-based chatbot on the website, we implemented a backend application using Flask and Python, deployed with Vercel. We used the Gemini-2.5-Flash model \cite{gemini25} for our chatbot because of our institution's license. 

\begin{figure}
\centering
\includegraphics[clip, trim=8.6cm 12.5cm 1.7cm 0cm, width=0.45\textwidth]{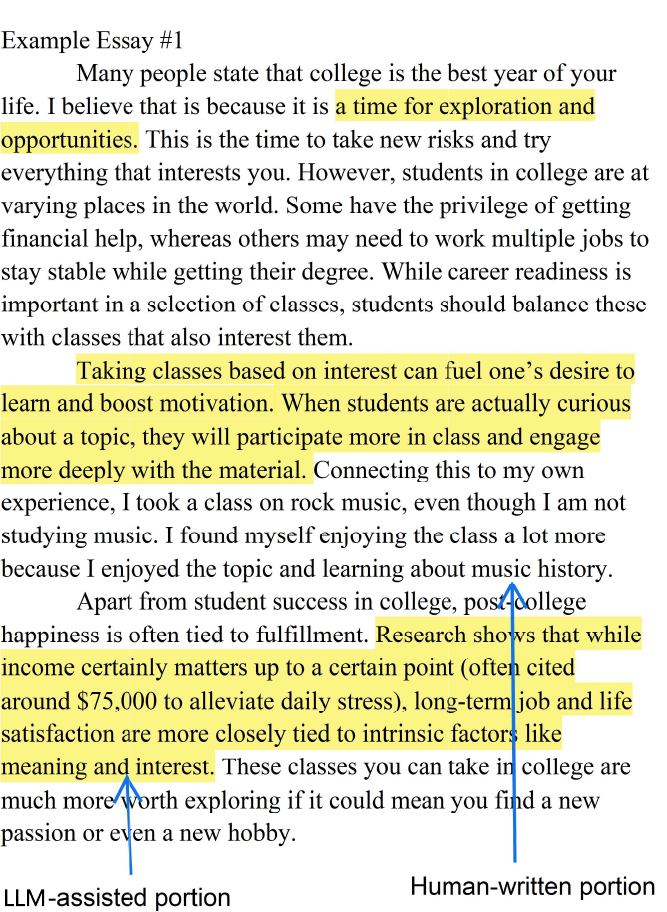}
\caption{Example annotated essay with highlighted LLM-assisted portions for the limited group. This student interweaves responses from LLM to support their main claim in this essay.}
\label{essay}
\end{figure}

Participants in the limited LLM access and unlimited LLM access groups interacted with their respective chatbots through a provided link. The writing assistant for the limited LLM group operated with a system prompt that restricted its responses to general guidance. The full system prompt is in Appendix \ref{sec:studylog}. This group was also given a maximum of three prompts to ask the LLM-based writing assistant, whereas the unlimited access group had no such limit. The number of prompts was selected due to the shortened length of the writing task. 
The limited LLM group had a maximum token output and a system prompt instruction to provide short responses that reflect general guidance and encourage students to develop their own ideas, rather than longer responses that directly answer the prompt. Meanwhile, the unlimited LLM group did not have a restricted token output or special instructions. 

Upon completion of the essay, participants were given a post-interaction survey to assess their perception of the writing process and their essay. Participants received a \$10 gift card as compensation.

\section{Methodology}
\subsection{Data Collection \& Annotation}
We collected the interactions with the LLM via our web application and annotated the writing actions and interactions. Due to the limited number of participants and short writing task given, we developed a simplified taxonomy inspired by previous research \cite{du-etal-2022-taxonomy, Nguyen03052024} with the following four writing intentions: writing, editing, structuring, and prompting.

We collected interaction data through a custom web application that logged participants' exchanges with the LLM, along with their writing activity in Google Docs. These logs enabled us to analyze both writing actions and prompting behavior throughout the essay-writing process.

To understand participants' writing behavior, we developed a taxonomy of four activities during the writing process: \textit{writing}, \textit{editing}, \textit{structuring}, and \textit{prompting}, informed by prior work on writing processes and revision behaviors. Writing refers to generating new text in full sentences, editing captures revisions to previously written content, and structuring involves organizing ideas or outlining the essay. Prompting means that the user is interacting with LLM to support writing. Two members of the research team manually annotated participants' writing intentions using this taxonomy, based on the Google Docs revision history and logged LLM interactions. Inter-annotator agreement for the writing intentions was assessed using Cohen’s $\kappa$, yielding substantial agreement ($\kappa = 0.71$).

\subsection{Metrics}\label{sec:metrics}
We established a set of metrics designed to capture the nuances of writing behavior, essay quality, lexical/semantic overlap with LLM output, and vocabulary richness, allowing us to correlate them with survey responses later in the analysis. Given our limited statistical power to detect all but very large effects on writing quality, we present descriptive statistics without asserting statistical significance.

\paragraph{Rubric-Based Evaluation}
Essays were evaluated using a custom rubric derived from standard English composition criteria, as shown in \ref{sec:appendix_essay}. The rubric assesses seven distinct dimensions: Focus and Details, Idea Support and Development, Organization, Voice, Vocabulary and Word Choice, Sentence Structure and Grammar, and Reads as human-written\footnote{This metric functions as a holistic AI-detection proxy rather than a compositional quality measure}, in which a higher score meant more human-like quality. Each category was scored on a five-point scale and accompanied by a brief justification. This rubric was inspired by standard English essay rubrics \cite{essay}.

To ensure reliability, grading was conducted independently by two university writing consultants employed at our institution's writing center, with the final rubric scores determined by averaging their two grades.\footnote{The graders were recruited via email and provided informed consent under an IRB-approved protocol.} 

\paragraph{Writing Time and Volume}
We tracked the total time each participant spent writing the essay (in minutes) to capture differences in writing effort and process efficiency across conditions. We also recorded the final word count of each essay as a measure of output length and productivity.

\paragraph{Lexical Diversity}
We measured lexical diversity using the Root Type-Token Ratio (Root TTR) \cite{guiraud1954-typetoken}. This metric provides a stable estimate of lexical diversity across essays of varying lengths by calculating the ratio of unique words (types) to the square root of the total words (tokens). Higher values indicate a broader, more diverse vocabulary usage relative to the total length of the text.

\paragraph{Prompt Type}
Prompts were categorized into four types: brainstorming/developing arguments, structure/organization, major revisions, and minor revisions. Developing arguments represented prompting to brainstorm talking points and specific facts or statistics. Structure/organization prompts focused on structuring the essay and making it cohesive. Major revisions included large revisions to the text, such as prompting to rewrite the entire essay, fixing grammatical mistakes, or modifying the essay stylistically. Lastly, minor revisions focused on rephrasing, sentence-level grammar checks, and asking for synonyms. Given the accuracy rate of annotations done by ChatGPT \cite{gpt-annotation}, and due to the large number of prompts, the prompts were initially organized using GPT-4 \cite{openai2024gpt4technicalreport} and all verified by one member of the research team with 94\% accuracy.

\paragraph{AI Similarity Scores}
To assess lexical overlap with LLM output and the degree to which participants incorporated LLM-generated text, we computed cosine similarity scores between the final student essays and the raw LLM responses generated during their specific writing session. Sentence embeddings were generated using the all-MiniLM-L6-v2 model \cite{reimers-2019-allmini}. This resulted in a similarity score ranging from 0 to 1, where higher scores indicate a greater direct incorporation of LLM-generated phrasing.

\section{Findings}
As we attempt to understand how different levels of LLM assistance influence writing outcomes and writing behaviors, we organize our analysis around these main research questions: [\textbf{RQ1}]: How does varied LLM assistance influence students’ writing behavior, including time allocation, writing strategies, and prompting practices? (Section \ref{sec:rq2}) [\textbf{RQ2}]: How do students perceive authorship, ownership, and acceptability of LLM-assisted writing? (Section \ref{sec:rq3})  [\textbf{RQ3}]: How does the level of LLM assistance affect writing quality, originality, and lexical diversity in student essays? (Section \ref{sec:rq1})

To address these questions, we combine rubric-based essay evaluation, lexical and similarity metrics, interaction logs, and survey responses. Each subsection below explicitly corresponds to one or more research questions.

\subsection{Writing Behavior and Interaction Patterns}\label{sec:rq2}
To examine how varied LLM assistance shapes writing behavior, we analyzed writing time, prompt usage patterns, and survey results across groups.

\begin{figure}[t]
\centering
\includegraphics[width=0.495\textwidth,trim=0 0 0 0cm, clip]{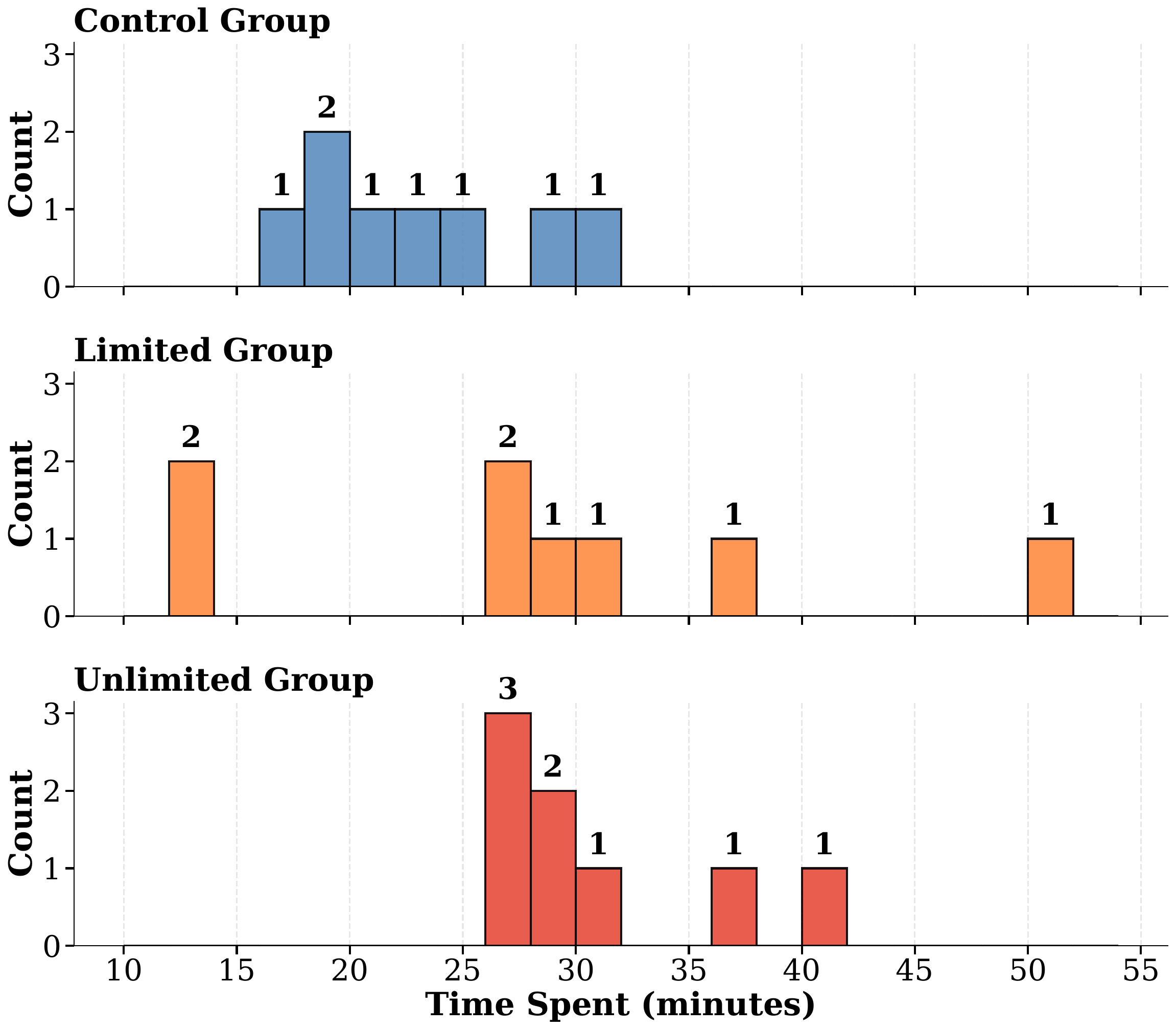}
\caption{Distribution of writing time (in minutes) by user group.}
\label{time}
\end{figure}

\paragraph{Writing Time}
Average writing time increased with greater LLM access. The control group completed essays in an average of 22.5 minutes, the limited LLM group in 27.8 minutes, and the unlimited LLM group in 30.4 minutes, as summarized in Table \ref{tab:core_metrics}. 
Notably, variance also differed across groups: the unlimited group showed the least spread in completion time (most participants clustered near the 30-minute mark), while the limited group showed the greatest variability (Figure \ref{time}).
One interpretation is that the three-prompt cap introduced individual differences in pacing: students who front-loaded prompts early finished faster, while those who saved prompts for revision extended their sessions.
The unlimited group, facing no strategic constraint, may have converged on a similar write-and-revise rhythm regardless of individual differences. No correlation between time and essay grade was observed.

\begin{figure}
\centering
\includegraphics[width=0.495\textwidth,trim=0 0 0 0cm, clip]{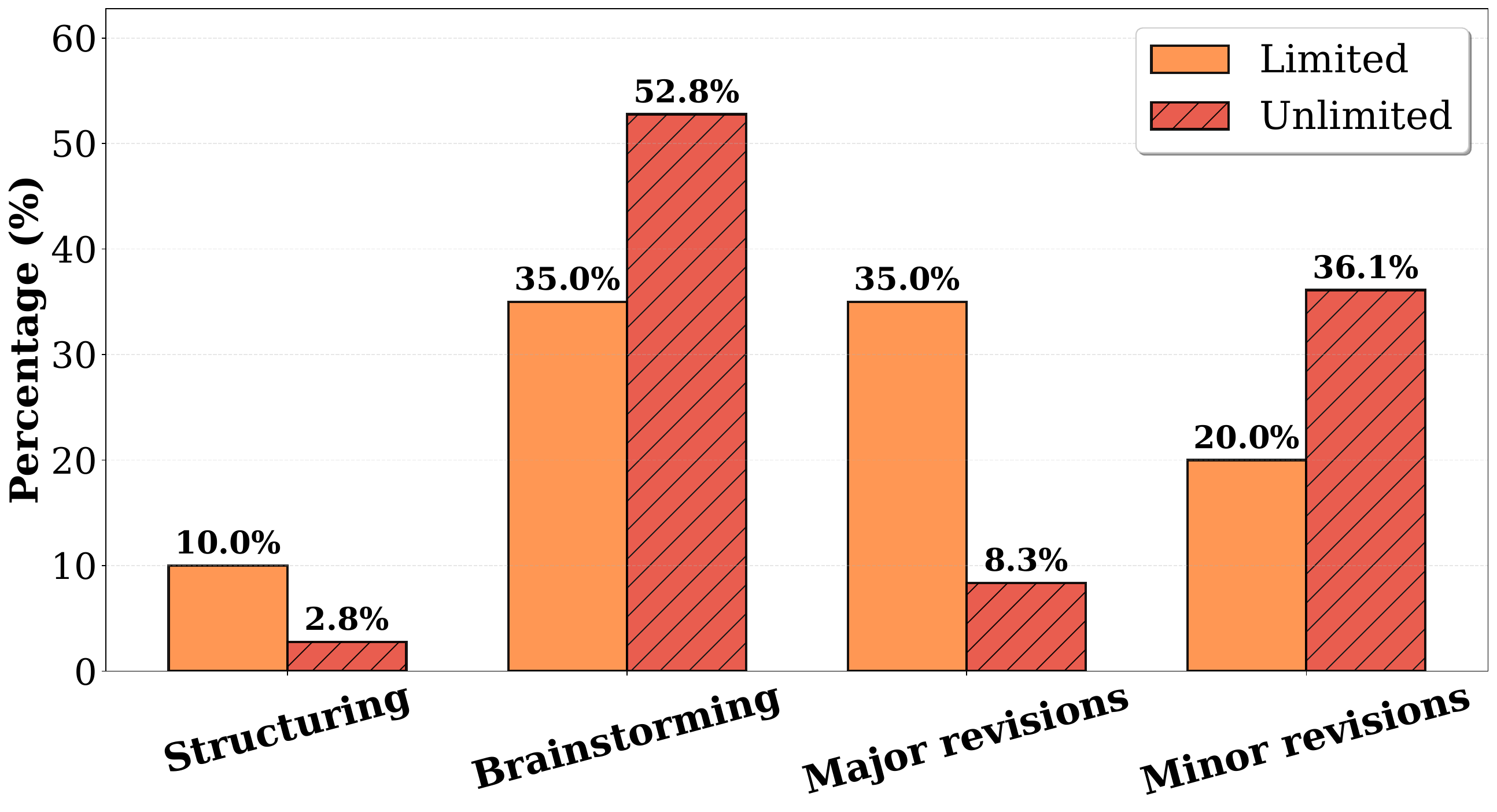}
\caption{Distribution of the type of prompts asked by the LLM groups.}
\label{prompttype}
\end{figure}

\begin{figure}[t]
\centering
\begin{subfigure}[t]{\columnwidth}
    \centering
    \includegraphics[width=\columnwidth,trim=0 0.3cm 0 0cm, clip]{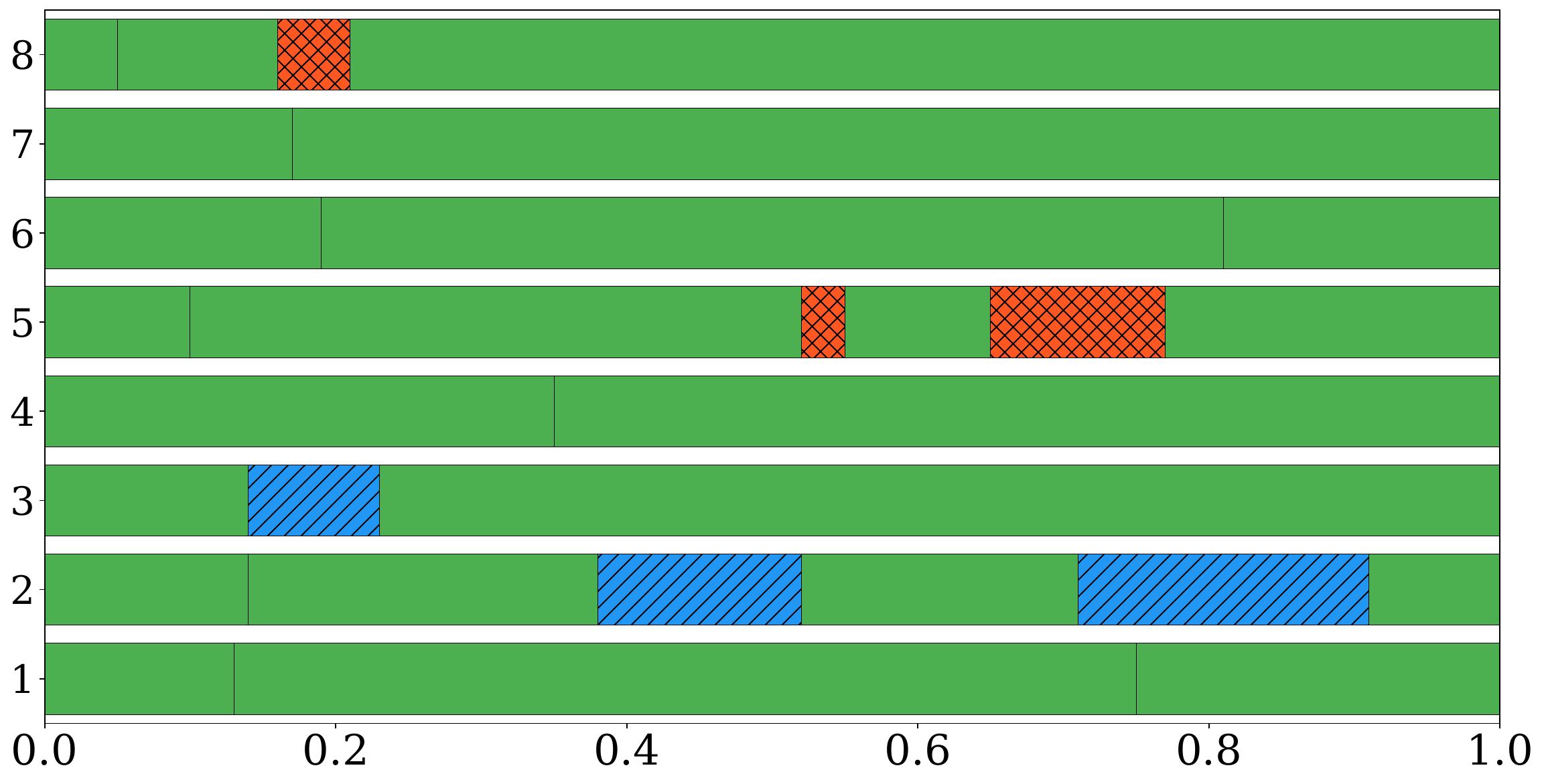}
    \caption{Control group}
    \label{gantt_control}
\end{subfigure}
\hfill
\begin{subfigure}[t]{\columnwidth}
    \centering
    \includegraphics[width=\columnwidth,trim=0 0.3cm 0 0cm, clip]{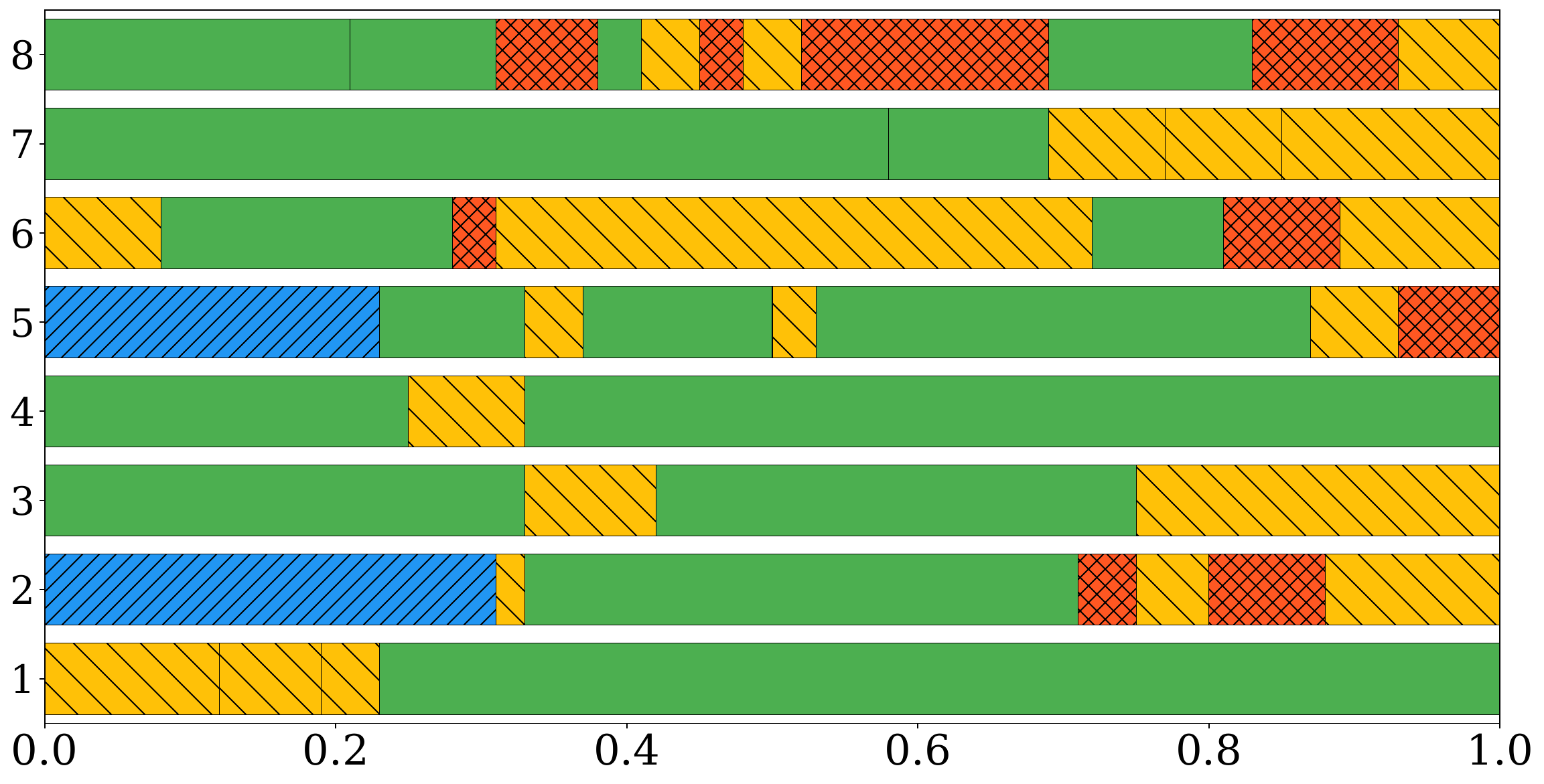}
    \caption{Limited LLM group}
    \label{gantt_limited}
\end{subfigure}
\hfill
\begin{subfigure}[t]{\columnwidth}
    \centering
    \includegraphics[width=\columnwidth,trim=0 0 0 0cm, clip]{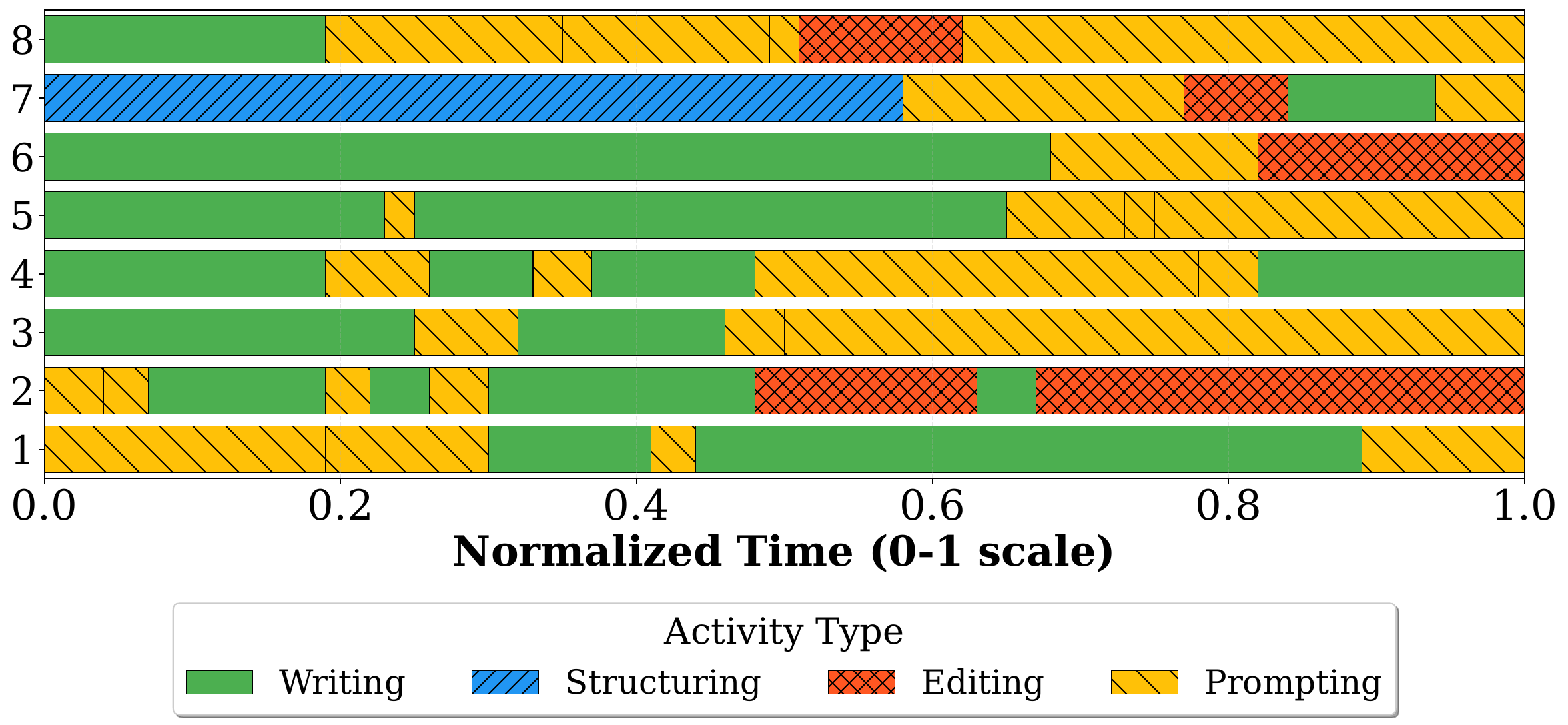}
    \caption{Unlimited LLM group}
    \label{gantt_unlimited}
\end{subfigure}

\caption{Distribution of writing intentions across the writing process for all user groups (8 participants per group). Each row represents a participant’s timeline, with colored segments indicating their activities across normalized time. 
}
\label{fig:gantt_all}
\end{figure}

\paragraph{Writing Process}
We analyzed temporal writing intentions by normalizing each participant’s writing session from start to finish. The limited group averaged 2.5 prompts, compared to 4.5 prompts for the unlimited group. Figures \ref{gantt_control}, \ref{gantt_limited}, and \ref{gantt_unlimited} illustrate how participants distributed effort across planning, writing, prompting, and editing.

As shown in Figure \ref{gantt_control}, two participants in the control group spent time on structuring the essay, and many participants spent most time writing sequentially, with additional time spent on editing the essay. Figure \ref{gantt_limited} shows the writing strategies of the limited LLM group. Six users demonstrated their collaboration with LLM, with writing themselves first, then prompting, and continuing this process until they were out of prompts. Despite having unlimited access to LLM, figure \ref{gantt_unlimited} shows that six of eight unlimited users still started the essay themselves, then prompted LLM for additional help. A majority of users in both groups with access to LLM utilized LLM in the latter part of their writing process (0.8-1). 

\textbf{Across all groups, most participants began by writing independently before seeking assistance}. Editing behaviors were equally prevalent across all groups, while explicit structuring activities were relatively rare across all conditions. The increased LLM access was not associated with a difference in the writing and prompting process. 

\paragraph{Prompt Types}
Figure \ref{prompttype} shows that brainstorming/argument development dominated prompts in both LLM assisted groups. Following argument-focused prompts, the unlimited group spent 36\% of their prompts on minor revisions to the text. The limited users distributed prompts differently, with 35\% on major revisions, and 20\% on minor revisions. 

Both groups showed low proportions of prompts directed toward structuring or organizing the essay. This pattern may reflect the possibility that participants may have completed structural planning independently, or the constrained length of the writing task may have reduced the concern over organization. The observed difference in major versus minor revision allocation between groups aligns with expectations that the restricted number of prompts encouraged the students to prioritize higher-impact revisions. 

\subsection{Perceptions of Ownership and Acceptability}\label{sec:rq3}
To understand students' perceptions of authorship when using LLM to help their writing process, we analyzed responses from post-interaction surveys. We report the behavioral metrics as descriptive statistics due to low statistical power.

\begin{table}[t]
\centering
\small
\setlength{\tabcolsep}{4pt}
\caption{Limited group correlation matrix between pre-survey and post-session measures ($*p < 0.05$).}
\label{tab:pearson_lim}
\begin{tabular}{lccc}
\toprule
& \textbf{Acceptance} & \textbf{Dev. Arguments} & \textbf{Time} \\
\midrule
\textbf{Crit. thinking eval.} & 0.92* & -- & -- \\
\textbf{Prompt literate}      & -0.80* & 0.71* & -- \\
\textbf{Relied on AI}         & -- & -- & 0.78* \\
\bottomrule
\end{tabular}
\end{table}

\begin{table}[t]
\centering
\small
\setlength{\tabcolsep}{4pt}
\caption{Unlimited group correlation matrix between pre-survey and post-session measures ($*p < 0.05$).}
\label{tab:pearson_unl}
\begin{tabular}{lccc}
\toprule
& \textbf{Acceptance} & \textbf{Similarity} & \textbf{Prompt \#} \\
\midrule
\textbf{Crit. thinking eval.} & -- & -0.79* & -- \\
\textbf{Saved time}           & -- & -- & -0.71* \\
\textbf{Comfortable with AI}  & -0.91* & -- & -- \\
\textbf{AI use frequency}         & -0.80* & -- & -- \\
\textbf{Difficulty}           & 0.81* & -- & -- \\
\bottomrule
\end{tabular}
\end{table}
\paragraph{Post-interaction survey}
Post interaction responses revealed ambivalence toward LLM-assisted authorship.

First, participants in the limited group expressed more ownership towards their essays. When asked if they would submit it as their own work, 62.5\% responded yes, 25\% were unsure, and 12.5\% declined. Most of the participants also perceived an increased ability to organize ideas (62.5\%). 

Correlational analysis within the limited group revealed several patterns (Table \ref{tab:pearson_lim}). Participants who used LLM frequently in writing assignments reported greater perceived need for critical thinking to decide what to accept or reject from the LLM ($r = 0.92, p < 0.05$). Those with self-reported prompt literacy accepted fewer prompts ($r = -0.80, p < 0.05$), but more frequently directed prompts towards developing arguments ($r = 0.71, p < 0.05$). This pattern suggests that the increased critical engagement with LLM outputs and prompting style may reflect an adjustment to the constrained nature of the system's responses compared to their previous exposure to LLM tools, such as ChatGPT. Participants who reported spending more time writing the essay also revealed feeling greater reliance on LLM ($r = 0.78, p < 0.05$), suggesting more time was spent on the LLM interface, prompting, and incorporating suggestions.

Conversely, \textbf{the unlimited group demonstrated less ownership over their work}. Only 25\% of these participants would submit their essay as their own work, and 75\% claimed unsure. This suggests that this group had increased concerns over academic integrity and perceived originality. However, alternative explanations, such as low confidence in content or structure, cannot be ruled out.

Correlational analysis for the unlimited group (\ref{tab:pearson_unl}) revealed several associations. A moderate negative correlation ($r = -0.79, p < 0.05$) between the cosine similarity score and the perceived critical thinking demands was observed. This implies that \textbf{students who actively evaluated and rejected LLM output produced more original work}, although causality cannot be determined. 

Students in the unlimited group who believed they saved time using LLM used fewer prompts ($r = -0.71, p < 0.05$). Conversely, comfort with LLM tools ($r = -0.91, p < 0.05$) and higher baseline LLM use frequency ($r = -0.80, p < 0.05$) were associated with lower prompt acceptance rates. Additionally, perceived task difficulty showed a positive relationship with prompt acceptance ($r = 0.81, p < 0.05$), suggesting that students reporting higher difficulty more readily incorporated suggestions.

Notably, across both groups with access to LLM, students with stronger self-reported prompting skill accepted fewer suggestions overall (limited: $r = -0.80$; unlimited: $r = -0.91$). This pattern could indicate that \textbf{students with stronger prompting literacy applied more critical judgment in evaluating suggestions}. These results support prior research that prompt-literate individuals are more likely to think more critically of the output and engage in iterative interactions with LLMs \cite{article}.

\textbf{Participants also expressed nuanced views on LLM use in coursework, aligning with previous studies \cite{aliakbari2026aiusage, Joshi_2025}}. Most responses indicated that LLM should be permitted for tasks such as brainstorming, source discovery, and concept clarification, but not for completing assignments or copying generated text. A small number of participants favored stricter limitations or unrestricted use, reflecting ongoing tension between perceived educational value and concerns about learning and academic integrity.

\subsection{Writing Quality and Originality} \label{sec:rq1}

To evaluate writing quality and originality, we analyzed essays using a rubric-based assessment, lexical diversity metrics, and similarity comparisons between student essays and LLM-generated responses.

\paragraph{Rubric Based Evaluation}\label{sec:rubric}
Table \ref{tab:core_metrics_acl} shows the average rubric scores, which were largely stable between groups. The statistical power of our study only allows us to detect very large effects of group on grading, and we do not find any significant differences between groups.

\begin{table}[t]
\centering
\small
\setlength{\tabcolsep}{6pt}
\caption{Average essay grades by user group, with highest scores in bold. Overall, differences between groups in grading outcomes are minor.}
\label{tab:core_metrics_acl}
\begin{tabular}{lccc}
\toprule
\textbf{Rubric Item} & \textbf{Control} & \textbf{Limited} & \textbf{Unlimited} \\
\midrule
Focus + Details     & 4.3  & 4.5  & \textbf{4.6} \\
Idea Support        & 4.6  & 4.5  & \textbf{4.7} \\
Organization        & 4.3  & 4.3  & \textbf{4.7} \\
Voice               & \textbf{4.9}  & 4.7  & 4.5 \\
Vocabulary          & 4.1  & \textbf{4.4}  & \textbf{4.4} \\
Sentence/Grammar    & 3.9  & \textbf{4.9}  & 4.7 \\
Reads as human-written & \textbf{4.9}  & 4.6  & 4.6 \\
\midrule
\textbf{Total Score} & 88.66\% & 88.39\% & 89.38\% \\
\bottomrule
\end{tabular}
\end{table}

\paragraph{Lexical Diversity}\label{sec:lexical}
As shown in Table \ref{tab:core_metrics}, the unlimited group exhibited the highest average Root TTR (9.555), followed by the limited group (9.4365) and the control group (8.963). These results suggest that \textbf{access to LLMs may encourage broader vocabulary use}. However, we again cannot definitively determine any statistical relationships given our limited power.

\paragraph{AI Similarity Scores}\label{sec:similarity}
The unlimited group exhibited a higher average similarity score (0.552) compared to the limited group (0.444), suggesting marginally greater incorporation of LLM output (See Table \ref{tab:core_metrics}). Although this is consistent with expectations, the difference was small and not statistically significant (ANOVA $p = 0.21$), preventing definitive conclusions regarding the dependence of LLM based only on similarity.

\begin{table}[t]
\centering
\small
\setlength{\tabcolsep}{6pt}
\caption{Average Metric Evaluations by User Group}
\label{tab:core_metrics}
\begin{tabular}{lccc}
\toprule
\textbf{Metric} & \textbf{Control} & \textbf{Limited} & \textbf{Unlimited} \\
\midrule
Essay Grade     & 88.66\%  & 88.39\%  & 89.38\% \\
Cosine Similarity        & --  & 0.444  & 0.552 \\
Root TTR        & 8.963  & 9.4365  & 9.555 \\
Writing Time (mins)   & 22.5  & 27.8  & 30.4 \\
Word Count          & 329  & 439  & 383 \\
\bottomrule
\end{tabular}
\end{table}

\section{Discussion}

Contrary to initial concerns, unlimited LLM access did not produce lower-quality essays than limited access or no LLM baselines. Only marginal differences in lexical diversity and similarity between LLM suggestions and essay text were exhibited. However, the unlimited group spent more time writing and exhibited a higher similarity to LLM output (0.552 vs. 0.444), suggesting greater LLM reliance or slower, more iterative integration of LLM text. This aligns with prior work on cognitive offloading, where increased AI reliance may reduce rather than increase self-directed cognitive engagement \cite{gerlich2025, kosmyna2025brainonchat}.

The LLM groups demonstrated varied prompting styles. Both groups exhibited human-AI collaboration through iterating, prompting, and editing the essay, but the limited group prioritized major revisions, whereas the unlimited group targeted minor mechanical revisions. This strategic reallocation may have been correlated with measurable ownership benefits. 62.5\% of limited participants reported willingness to submit their essay as independent work, compared to 25\% in the unlimited group. Similarly, 62.5\% of the limited group reported improved organizational ability.

Correlation analysis between pre- and post-interaction survey responses exhibited differences in how users interacted with the AI. As shown in Table \ref{tab:pearson_unl}, limited group participants who are comfortable with AI tools reported elevated critical thinking demands ($r = 0.92, p < 0.01$) and focused prompts on argument-development help ($r = 0.71, p < 0.05$). This pattern suggests that AI-literate students were more cautious and critical of the output. The unlimited LLM group exhibited a strong negative correlation between perceived need for critical thinking and LLM similarity scores. This suggests that students who evaluated LLM responses more rigorously produced more original work. 

While constrained LLM access appeared to encourage strategic use and increase feelings of authorship, the study’s small sample size and simplified writing task limit generalizability. 

\section{Conclusion}
This study examined how varied levels of LLM assistance influence college students’ writing quality, writing behavior, and perceptions of authorship. In an IRB-approved study with 24 participants, results showed that overall essay quality showed no statistical significance, and systematic patterns in writing behavior, cognitive engagement, and ownership views revealed nuances about how LLMs iimpactstudent writing. 

These findings motivate future research on LLM-assisted writing in more realistic academic settings, with larger samples and extended writing tasks, to better understand how students balance AI support with independent authorship.

\section{Limitations}
This study has several limitations. The data were collected over a two-week period from native English speaking college students in authors' university, which limits generalizability despite efforts to recruit participants from diverse backgrounds. The sample size was small, with 24 participants, due to time and resource constraints. As a result, the study focused on a single opinion-based essay prompt, which may not reflect writing behaviors or LLM interactions in other contexts, such as research-driven or evidence-based assignments. Longer writing tasks may reveal different patterns. Future work should explicitly examine EFL/ESL learners, who represent a significant and potentially more LLM-reliant group.

The model also lacked built-in conversational memory. Participants assumed that the LLM would retain context across prompts, leading to confusion when follow-up prompts were interpreted in isolation. This issue was only identified after receiving feedback from the initial participants. 

Additional limitations arose in data collection and annotation. Google Docs editing histories only captured major revisions, preventing fine-grained analysis of sentence-level edits and micro revisions during continuous writing. Manual annotation of writing intentions, while validated through inter-rater agreement ($\kappa = 0.71$), introduced potential bias. Additionally, we used Gemini-2.5-Flash rather than ChatGPT, which may limit comparison to students' real-world AI experiences. We did not observe any clear attempts to circumvent the system prompt constraints. Alternative explanations for correlational outcomes (e.g., low confidence rather than AI-induced ambiguity) cannot be fully ruled out.

We also used cosine similarity to determine an AI similarity score, which captures surface-level semantic overlap but does not directly measure originality in a deeper cognitive sense. 

Future work should expand participant pools across institutions and regions, include multiple writing tasks of varying genres and lengths, incorporate memory-enabled frameworks to better reflect real-world AI writing tools, and employ more detailed logging mechanisms to capture granular writing behavior. Longitudinal designs examining skill development would also help capture long term effects of varied LLM access on writing ability.

\section{Ethical Considerations}
This study does have ethical considerations related to participant privacy, informed consent, and responsible use of generative AI. All participants were fully informed of the purpose of this study and the structure of the writing tasks. Participation was voluntary, and students were free to withdraw at any time without penalty. Incentives were given to participants as per gift cards in return for their time and effort. 
The study also considered the ethical implications of exposing participants to biased, inaccurate, or misleading AI-assisted output. The LLM was constrained to provide short, general writing suggestions rather than fully generated paragraphs or factual claims. Our study does have potential harm if there is a misinterpretation of our findings, which could encourage the misuse or altogether elimination of AI tools in a school environment. To address this, we emphasized the balance, acknowledgment, and responsibility of AI use, especially in an academic setting.

\bibliography{custom, anthology}

\clearpage
\appendix
\label{sec:Appendix}

\section{System Design Details}
\subsection{Study Procedure}
\label{sec:Appendix_procedure}
Participants were recruited via email and asked to complete a preliminary survey administered through Google Forms. Upon receipt of the completed survey, participants were scheduled for an individual, in-person study session.

Each session took place in a private room. Participants were first asked to read and sign the informed consent form (Appendix~\ref{subsubsec:consent}) before beginning the study. Following consent, participants received an email containing two links: one to a document outlining study instructions and another to a Google Doc for composing their essay. Participants assigned to the limited-AI and unlimited-AI conditions additionally received login credentials via email to access the AI-assisted writing platform. The full set of condition-specific instructions is provided in Appendix~\ref{subsubsec:instruction}.

All participant data were stored in a private Google Drive folder accessible only to members of the research team. Identifying information was removed from all essays and replaced with anonymized user IDs. During data collection through the LLM interface, only anonymized user IDs were recorded to protect participant privacy.

\subsubsection{Consent Form}
\label{subsubsec:consent}
You are invited to participate in a research project investigating the Overdose Effect of AI-Assisted Writing at the University of Minnesota - Twin Cities. We ask that you read this form and ask any questions you may have before agreeing to take part.

\noindent
\textbf{Purpose of the Study:} This project explores how different levels of reliance on LLM tools (such as ChatGPT or other generative AI systems) affect students’ writing ability, creativity, and critical thinking. The goal is to understand how LLM assistance may influence the quality, style, and originality of written work.

\noindent
\textbf{Procedures:} The participant will engage in a short preliminary survey asking about their background, experience with AI tools, and comfort level using them. We will then have the participant write a short essay (at least 250-300 words) in response to a provided prompt within approximately 30 minutes. Lastly, the participant will be asked to complete a short post-survey reflecting on their writing process and any use of AI during the activity.

\noindent
\textbf{Risks and Benefits of Being in the Study:} There are minimal risks to being in the study. While there is no direct benefit to you, this study will help us better understand how AI tools impact writing skills and academic performance. Each user will be compensated with a \$10 gift card for their time.

\noindent
\textbf{Data Sharing:} The information you provide through this project will be recorded.  Data that you provide as part of this research project will be shared with the other students on the project team, and the course staff.

\noindent
\textbf{Confidentiality:} We will not collect any information that will make it possible to trace your participation back to you. We will not share your participation with anybody outside of the project team. We will keep your participation private to the extent allowable by law.

\noindent
\textbf{Voluntary Nature of Research Participation:} Participation in this project is voluntary. Your decision whether or not to participate will not affect your current or future relations with the University of Minnesota - Twin Cities.

\noindent
\textbf{Contacts and Questions:} The students conducting this research project are: Julia Christenson, Shirley Hayati, and Karin de Langis. The instructor supervising this project is Dongyeop Kang. You may ask any questions you have now. If you have questions later, you are encouraged to contact us. You may keep this page for your records after signing and returning the attached sheet.

\noindent
\textbf{Statement of Consent:} I have read the attached information regarding the project investigation of the Overdose Effect of AI-Assisted Writing at the University of Minnesota - Twin Cities. I have asked questions and have received answers. I consent to participate in this project.

\subsubsection{Instruction Forms}
\label{subsubsec:instruction}

\noindent
\textbf{Control Group}
"A Google Doc will have been shared with you for you to write your essay in. Write a response to the prompt below. Your response should be around 300 words (at least 3 paragraphs), however, you can write more if you want to. Take as much time as you need to write your response. 

Prompt:
Write a response in which you discuss the extent to which you agree or disagree with the recommendation and explain your reasoning for the position you take. \textit{College students should be encouraged to pursue subjects that interest them rather than the courses that seem most likely to lead to jobs. }

Once you are done writing your response fill out the following survey. Once you are finished with the essay and the survey please let one of the study team members know."

\noindent
\textbf{Limited Group} 
"A Google Doc will have been shared with you for you to write your essay in. Write a response to the prompt below. Your response should be around 300 words (at least 3 paragraphs), however, you can write more if you want to. Take as much time as you need to write your response. 

When writing your response, \textbf{you are allowed to prompt AI 3 times}. You don’t have to use all or any of your allotted prompts; however, you cannot use more than 3. You also cannot prompt the AI to write the full response for you.
To access AI, go to this link and sign in with the username and password that the study team member gave you.

Prompt:
Write a response in which you discuss the extent to which you agree or disagree with the recommendation and explain your reasoning for the position you take. \textit{College students should be encouraged to pursue subjects that interest them rather than the courses that seem most likely to lead to jobs. }

Once you are done writing your response fill out the following survey. Once you are finished with the essay and the survey please let one of the study team members know."

\noindent
\textbf{Unlimited Group}
"A Google Doc will have been shared with you for you to write your essay in. Write a response to the prompt below. Your response should be around 300 words (at least 3 paragraphs), however, you can write more if you want to. Take as much time as you need to write your response. 

When writing your response, \textbf{you are allowed to prompt AI as many times as you would like} to help you write your response. However, you cannot prompt the AI to write the full response for you. (You cannot copy and paste in the prompt.)
To access AI, go to this link and sign in with the username and password that the study team member gave you.

Prompt:
Write a response in which you discuss the extent to which you agree or disagree with the recommendation and explain your reasoning for the position you take. \textit{College students should be encouraged to pursue subjects that interest them rather than the courses that seem most likely to lead to jobs. }

Once you are done writing your response fill out the following survey. Once you are finished with the essay and the survey please let one of the study team members know."

\subsection{Study Interface}
\label{sec:studylog}
The full set of instructions for the limited AI model before the user's prompt is below.

\begin{quote}
\texttt{SYSTEM\_PROMPT = "You are a helpful study assistant. Your role is to provide general guidance, suggestions, and help students think through their work.
IMPORTANT RULES:
- Keep ALL responses under 100 words - be concise and focused
- Provide suggestions, outlines, and general guidance only
- DO NOT write complete essays, paragraphs, or full answers that students can copy-paste
- Help students develop their own ideas through questions and prompts
- Offer structural advice (e.g., "Consider organizing your essay with: intro, 3 body paragraphs, conclusion")
- Suggest topics to research or think about
- Ask clarifying questions to help students think deeper
If asked to write a complete essay or full answer, politely decline and instead offer to:
1. Help them brainstorm ideas
2. Suggest an outline structure
3. Provide tips on how to approach the topic
4. Ask questions to help them develop their thoughts"}
\end{quote}

The interactions with the chatbot and user were sent to a csv file using Vercel Blob. Each row of data followed this order: timestamp, user\_ID, user\_group, prompt\_num, prompt, response. 

\section*{Preliminary Survey}

\subsection*{A. Demographics \& Background}
\noindent Below are the questions included in the preliminary survey instrument.
\begin{enumerate}[label=Q.\arabic*]
    \item Name (first last): \dotfill
    
    \item Email: \dotfill
    
    \item Age: \dotfill
    
    \item Ethnicity: \dotfill
    
    \item Year:
          \begin{enumerate}[label=(\alph*), noitemsep, topsep=0pt]
              \item Freshman
              \item Sophomore
              \item Junior
              \item Senior
              \item Grad student
              \item Other: \dotfill
          \end{enumerate}
    
    \item Gender: \dotfill
    
    \item Major/minor/field of study: \dotfill
    
    \item Category of field of study: \dotfill
    
    \item Are you a native English speaker?
          \begin{enumerate}[label=(\alph*), noitemsep, topsep=0pt]
              \item Yes
              \item No
          \end{enumerate}
    
    \item Which generative AI tools have you used before? (Select all that apply)
          \begin{enumerate}[label=(\alph*), noitemsep, topsep=0pt]
              \item ChatGPT
              \item Bard
              \item Claude
              \item Co-pilot
              \item Built-in tools (Grammarly, Word, etc.)
              \item Other: \dotfill
          \end{enumerate}
    
    \item How often do you currently use AI tools for writing tasks? \\
          (1=Never) \quad 1 \quad 2 \quad 3 \quad 4 \quad 5 \quad (5=Always)
\end{enumerate}

\subsection*{B. Self-Assessments}
\noindent Rate your agreement with the following statements (1 = Strongly disagree, 5 = Strongly agree).
\begin{enumerate}[label=Q.\arabic*, resume]
    \item I consider myself a strong writer. \dotfill ( )
    
    \item I think AI will improve my writing ability. \dotfill ( )
    
    \item I worry that using AI will make me less capable as a writer. \dotfill ( )
    
    \item I am comfortable using AI tools for drafting, revising, or editing. \dotfill ( )
    
    \item I understand how to prompt AI (I know how to give clear prompts). \dotfill ( )
    
    \item Of your last 10 course-writing tasks, how many used AI in any way? (0--10): \dotfill
\end{enumerate}

\subsection*{C. Behavioral Baseline \& Motivations}
\begin{enumerate}[label=Q.\arabic*, resume]
    \item Why do you use or avoid AI tools for writing? (1--2 sentence open text) \\
          \dotfill \\
          \dotfill \\
          \dotfill
    
    \item Any comments, questions, concerns? \\
          \dotfill \\
          \dotfill \\
          \dotfill
\end{enumerate}

\newpage

\section*{Post-Interaction Survey}

\subsection*{A. Usage Log (Only asked to AI groups)}
\noindent Below are the questions included in the post-interaction survey instrument.
\begin{enumerate}[label=Q.\arabic*]
    \item Approximately what percent of your final text was directly generated or suggested by AI?
          \begin{enumerate}[label=(\alph*), noitemsep, topsep=0pt]
              \item 0\%
              \item 1--10\%
              \item 11--25\%
              \item 26--50\%
              \item 51--75\%
              \item 76--99\%
              \item 100\%
          \end{enumerate}
    
    \item Did you copy-paste any AI output into your final submission?
          \begin{enumerate}[label=(\alph*), noitemsep, topsep=0pt]
              \item Yes
              \item No
          \end{enumerate}
    
    \item Did you edit the AI output?
          \begin{enumerate}[label=(\alph*), noitemsep, topsep=0pt]
              \item No edits
              \item Minor edits (wording)
              \item Moderate edits (structure)
              \item Major edits (rewrote most)
          \end{enumerate}
\end{enumerate}

\subsection*{B. Process \& Experience}
\noindent Rate your agreement with the following statements (1 = Strongly disagree, 5 = Strongly agree).
\begin{enumerate}[label=Q.\arabic*, resume]
    \item The AI suggestions helped me generate ideas. \dotfill ( )
    
    \item The AI suggestions saved me time. \dotfill ( )
    
    \item The AI suggestions improved the clarity of my writing. \dotfill ( )
    
    \item The AI suggestions reduced my creative expression. \dotfill ( )
    
    \item I felt I relied too heavily on AI to complete the task. \dotfill ( )
    
    \item I had to do substantial critical thinking to decide what to accept or reject from the AI. \dotfill ( )
    
    \item I felt the final product still reflected my own voice. \dotfill ( )
    
    \item I would submit the final essay as my own in a regular class.
          \begin{enumerate}[label=(\alph*), noitemsep, topsep=0pt]
              \item Yes
              \item No
              \item Not sure
          \end{enumerate}
    
    \item I would disclose to an instructor that I used AI for this piece.
          \begin{enumerate}[label=(\alph*), noitemsep, topsep=0pt]
              \item Yes
              \item No
              \item Maybe
          \end{enumerate}
\end{enumerate}

\subsection*{C. Cognitive Load / Effort \& Time}
\begin{enumerate}[label=Q.\arabic*, resume]
    \item How difficult was it to complete this task? \\
          (Very easy) \quad 1 \quad 2 \quad 3 \quad 4 \quad 5 \quad (Very difficult)
    
    \item How much of the 30 minutes did you spend editing vs. composing vs. reading AI outputs? \\
          Editing: \dotfill \% \\
          Composing: \dotfill \% \\
          Reading AI outputs: \dotfill \%
\end{enumerate}

\subsection*{D. Reflection on Skills}
\noindent After this session, indicate how you feel your ability changed in each area (increase / no change / decrease).
\begin{enumerate}[label=Q.\arabic*, resume]
    \item Organize ideas: \dotfill
    
    \item Use precise vocabulary: \dotfill
    
    \item Write creatively: \dotfill
    
    \item Perform critical analysis: \dotfill
\end{enumerate}

\subsection*{E. Originality \& Authenticity}
\begin{enumerate}[label=Q.\arabic*, resume]
    \item How original do you feel your final essay is? \\
          (Not at all original) \quad 1 \quad 2 \quad 3 \quad 4 \quad 5 \quad (Very original)
    
    \item On a scale of 0--100, how authentic does the essay feel to you? \dotfill
    
    \item Would you be comfortable if this exact essay were used as a graded assignment?
          \begin{enumerate}[label=(\alph*), noitemsep, topsep=0pt]
              \item Yes
              \item No
              \item Maybe
          \end{enumerate}
\end{enumerate}

\subsection*{F. Ethics \& Academic Integrity}
\begin{enumerate}[label=Q.\arabic*, resume]
    \item Do you think using AI in coursework should be allowed?
          \begin{enumerate}[label=(\alph*), noitemsep, topsep=0pt]
              \item Yes
              \item No
              \item Depends --- please explain: \dotfill
          \end{enumerate}
    
    \item If allowed, should students be required to disclose AI usage?
          \begin{enumerate}[label=(\alph*), noitemsep, topsep=0pt]
              \item Yes
              \item No
              \item Depends
          \end{enumerate}
\end{enumerate}

\subsection*{G. Open Reflection}
\begin{enumerate}[label=Q.\arabic*, resume]
    \item Describe (in 1--3 sentences) how you used AI in this task (what prompts you gave, what you accepted/rejected). \\
          \dotfill \\
          \dotfill \\
          \dotfill
    
    \item What, if anything, did AI prevent you from doing or thinking about? \\
          \dotfill \\
          \dotfill \\
          \dotfill
    
    \item Did anything about the AI outputs surprise you?
          \begin{enumerate}[label=(\alph*), noitemsep, topsep=0pt]
              \item Yes --- please explain: \dotfill
              \item No
          \end{enumerate}
    
    \item Any additional comments about how AI influenced your thinking or writing on this task? \\
          \dotfill \\
          \dotfill \\
          \dotfill
\end{enumerate}

\subsection*{H. Optional Detection / Guessing Task}
\begin{enumerate}[label=Q.\arabic*, resume]
    \item We sometimes run automatic classifiers on texts. Do you think your submitted essay was more likely:
          \begin{enumerate}[label=(\alph*), noitemsep, topsep=0pt]
              \item Human-written
              \item AI-assisted
              \item Mostly AI-generated
          \end{enumerate}
\end{enumerate}

\section*{Essay Grading Rubric}
\label{sec:appendix_essay}
\noindent The following rubric was used to evaluate the quality of student essays across all experimental conditions. Each category is scored on a scale of 1--5 points.

\begin{table*}[t]
\centering
\scriptsize
\setlength{\tabcolsep}{3pt}
\begin{tabular}{|p{1.8cm}|p{2.3cm}|p{2.3cm}|p{2.3cm}|p{2.3cm}|p{2.3cm}|}
\hline
\textbf{Category} & \textbf{5} & \textbf{4} & \textbf{3} & \textbf{2} & \textbf{1} \\
\hline
\textbf{Focus \& Details} & 
Engaging and full development of a clear thesis as appropriate to the assignment purpose. & 
Competent and well-developed thesis; thesis represents a sound and adequate understanding of the assigned topic. & 
Mostly intelligible ideas; thesis is weak, unclear, too broad, or only indirectly supported. & 
Most simplistic and unfocused ideas; little or no sense of purpose. & 
Ideas are extremely simplistic, showing signs of confusion and misunderstanding of the prompt. The thesis is missing or not discernible. \\
\hline
\textbf{Ideas, Support, \& Development (Evidence)} & 
Consistent evidence with originality and depth of ideas. Ideas work together, and the main points are sufficiently supported with evidence. & 
Ideas are supported sufficiently; support is sound, valid, and logical. & 
Main points and ideas are only indirectly supported. Support is loosely relevant to the main points. & 
Insufficient or non-specific support. & 
Lack of support for main points. \\
\hline
\textbf{Organization} & 
Organization is sequential and appropriate. Paragraphs are well developed and appropriately divided. & 
Competent organization, without sophistication. May lack effective transitions. & 
Limited attempts to organize around a thesis; paragraphs are mostly stand-alones with weak transitions. & 
Organization, while attempted, was unsuccessful. Paragraphs were simple, disconnected, and formulaic. & 
Little to no organization; disjointed or confusing structure. \\
\hline
\textbf{Voice} & 
The author's purpose is very clear. Their extensive knowledge and/or experience with the topic is/are evident. & 
The author's purpose is mostly clear. Some evidence of attention to the audience. The author's knowledge and/or experience with the topic is/are evident. & 
Little or inconsistent sense of the audience related to the assignment purpose. Tone and point of view are not refined or consistent. & 
Shows almost no awareness of a particular audience; reveals no grasp of appropriate tone and/or point of view for given assignment. & 
Lacks awareness of a particular audience for the assignment; tone and point of view are somewhat inappropriate or very inconsistent. \\
\hline
\textbf{Vocabulary and Word Choice} & 
Exceptional vocabulary range, accuracy, and correct and effective word usage. & 
Good vocabulary range and accuracy of usage. & 
Ordinary vocabulary range, mostly accurate. & 
Errors of diction and usage, while evident, do not interfere with readability. & 
Extremely limited vocabulary; choices lack grasp of diction. \\
\hline
\textbf{Sentence Structure, Grammar, Mechanics, \& Spelling} & 
All sentences are well constructed and have varied structure and length. The author makes no errors in grammar, mechanics, and/or spelling. & 
Most sentences are well constructed and have varied structure and length. Contains only occasional punctuation, spelling, and/or capitalization errors. & 
Sentences show a formulaic or tedious sentence pattern. Some errors in grammar, mechanics, and/or spelling, but they do not interfere with understanding. & 
Most sentences are well constructed, but they have a similar structure and/or length. The author makes several errors in grammar, mechanics, and/or spelling that interfere with understanding. & 
Sentences sound distractingly repetitive or are difficult to understand. The author makes numerous errors in grammar, mechanics, and/or spelling that interfere with understanding. \\
\hline
\textbf{AI Detection/ Human-written} & 
I am confident that this text was completely human-written. This essay feels completely authentic and original. & 
I am confident that this text was written mostly by a human. This essay feels authentic, but may have had some AI assistance. & 
I believe this essay was done by a human with AI assistance. Some portions of text seem to be AI-generated. & 
I believe this essay was mostly written with the help of AI. A large portion of the text sounds repetitive and/or unoriginal. & 
I believe this essay was completely written by AI. \\
\hline
\end{tabular}
\end{table*}

\vspace{1em}
\noindent \textbf{Scoring:} Each category is scored from 1--5 points. Total possible score for the first six categories: 30 points.
\medskip

\end{document}